# THE COGNITIVE PROCESSING OF CAUSAL KNOWLEDGE

Scott B. Morris
Institute of Psychology
Illinois Institute of Technology
Chicago, Il 60616
morris@charlie.cns.iit.edu

Doug Cork
Department of Biology
Illinois Institute of Technology
Chicago, Illinois 60616
cork@charlie.cns.iit.edu

Richard E. Neapolitan
Computer Science Department
Northeastern Illinois University
5500 N. St. Louis
Chicago, Illinois 60625
R-Neapoltan@neiu.edu

## ABSTRACT

There is a brief description of the probabilistic causal graph model for representing, reasoning with, and learning causal structure using Bayesian networks. It is then argued that this model is closely related to how humans reason with and learn causal structure. It is shown that studies in psychology on discounting (reasoning concerning how the presence of one cause of an effect makes another cause less probable) support the hypothesis that humans reach the same judgments as algorithms for doing inference in Bayesian networks. Next, it is shown how studies by Piaget indicate that humans learn causal structure by observing the same independencies and dependencies as those used by certain algorithms for learning the structure of a Bayesian network. Based on this indication, a subjective definition of causality is forwarded. Finally, methods for further testing the accuracy of these claims are discussed.

## 1 INTRODUCTION

A new perspective on causation has emerged from researchers in artificial intelligence. In efforts to create systems that can reason with causal relationship, they developed a probabilistic graphical model of causality. The probabilistic graphical structure for representing causal relationship is called a **Bayesian network**. The combination of representing, reasoning with, and learning causal structure using Bayesian networks we term here as the **Probabilistic Causal Graph (PCG) Model**. This model has proved useful in artificial intelligence and expert systems applications. However, is it related to how humans reason with and learn causal knowledge?

The importance of understanding human reasoning to artificial intelligence does need elaboration. Judea Pearl [1986, 1995] has long argued that humans perform inference with existing causal knowledge in the same way as a well-known algorithm for doing inference in a Bayesian network. In Section 3.1 we summarize his argument and cite research on human subjects that support it. Then, in Section 3.2 we present the main result of this paper. That is, we hypothesize that humans learn causal knowledge by observing the same independencies and dependencies used by certain algorithms for learning the structure of a Bayesian network. We support this claim with results of studies by Piaget [1952, 1954, 1966] on infants and children. Our conjecture, together with Pearl's, constitute a model of how humans reason with and learn causes. We begin in Section 2 with a brief description of the probabilistic causal graphical model.

## 2 THE PCG MODEL

The Probabilistic Causal Graph Model assumes that the causal relationships among a set of variables $\mathcal{U}$ can be modeled by a directed acyclic graph (DAG) $\mathcal{D}$ (called a Bayesian network) in which each node consists of an element of $\mathcal{U}$, and a joint probability distribution $\mathcal{P}$, on the variables in $\mathcal{U}$, which satisfies the *Markov* and *faithfulness* conditions for $\mathcal{D}$. The edges in $\mathcal{D}$ are meant to represent direct causal influences. Figure 1 shows a Bayesian network in which the variables have to do with the causal mechanisms underlying how pavement could get wet. These are probabilistic rather than deterministic relationships. For example, the pavement will not get wet even when the sprinkler is on if the pavement is covered with a blanket. Notice there is no edge from $K$ to $L$ because the sprinkler does not directly cause a slippery pavement. Rather it is only through making it wet that this happens. There are only edges from direct causes to effects. In general, if we found out the sprinkler were on, it should increase the probability that the pavement would be slippery because it would make it more probable that the pavement were wet. However, according to this model, if we already knew the pavement was wet, there would be a fixed probability it was slippery. Finding out the sprinkler was on would not increase that probability. The idea is that knowledge of an effect's direct causes shields the effect from the influence of variables that can affect those causes. This is the Markov condition, which is stated formally as follows. A probability distribution $\mathcal{P}$ on the variables in $\mathcal{D}$ satisfies the **Markov condition** for $\mathcal{D}$ if the value assumed by a variable $X$ in $\mathcal{D}$ is probabilistically independent of the values of all other variables in $\mathcal{D}$, except the descendents of $X$, conditional on the parents of $X$.

Next we discuss the faithfulness condition. Consider again Figure 1. Intuitively, the sprinkler being on should increase the probability of it being summer, which should therefore decrease the probability it had rained. The Markov condition only requires that the sprinkler and rain be independent given the season; it does not require that they be dependent if we do not know the season. Here is where the faithfulness condition comes in. A probability distribution $\mathcal{P}$ on the variables in $\mathcal{D}$ satisfies the **faithfulness condition** for $\mathcal{D}$ if all the conditional independencies true in $\mathcal{P}$ are entailed by the Markov condition applied to $D$.

The Probabilistic Causal Graph Model includes algorithms for reasoning with and learning causes. The next section discusses ones most pertinent to our research plans. Summaries of other algorithms are in [Pearl, 1988], [Neapolitan, 1990], [Spirtes et al, 1993], [Heckerman et al, 1994], [Glymour, 1996], and [Castillo et al, 1997].

# 3 THE PCG MODEL AND HUMAN REASONING

The Probabilistic Causal Graph Model does much to refute Christensen's [1990, p.279] claim that 'causation is not something that can be established by data analysis. Establishing causation requires logical arguments that go beyond the realm of numerical manipulation.' However, does it have anything to do with how humans reason with and learn causes? We discuss arguments for each of these next.

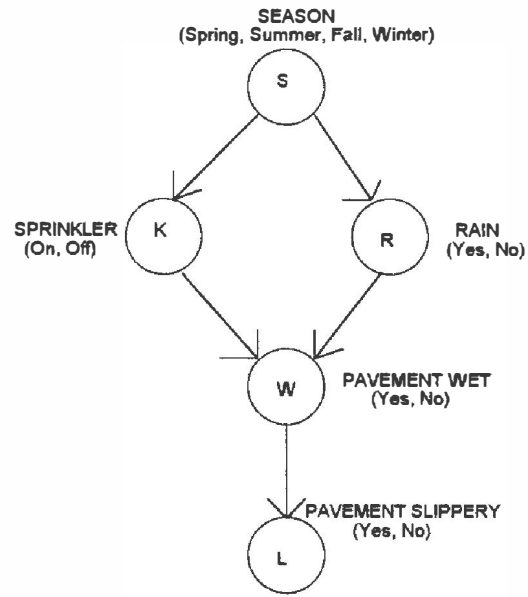

**Figure 1 – A Bayesian Network**

## 3.1 REASONING WITH CAUSES

Pearl [1988, 1995] argues that humans structure causal knowledge in a Bayesian network and reason with that knowledge similar to a message-passing algorithm developed in [Pearl, 1986]. Pearl's argument is not that a globally consistent Bayesian network exists at the cognitive level. 'Instead, fragmented structures of causal organizations are constantly being assembled on the fly, as needed, from a stock of functional building blocks' - [Pearl, 1995]. For example, suppose a Mr. Holmes has observed earthquakes trigger his alarm in the past. Having learned that his alarm had sounded at home, he would assemble the cause-effect edge directed from BURGLAR to ALARM. He would reason along this edge in the direction from ALARM to BURGLAR to conclude that he had probably been burglarized. If he later learned of an earthquake, he would assemble the EARTHQUAKE→ALARM directed edge. He would then reason that the earthquake explains away the alarm, and therefore he had probably not been burglarized. If, when Mr. Holmes got home, he saw strange footprints in the yard, he would assemble the BURGLAR→FOOTPRINTS directed edge and reason along that edge. This tracing of edges is how Pearl's [1986] algorithm for doing inference in a Bayesian network proceeds. That algorithm updates probabilities in sequence along the edges using Bayes' Rule.

Recent research in psychology lends some support to the argument that humans reach the same judgments as those that would be reached by algorithms for doing in-ference in a Bayesian network. Psychologists have long been interested in how an individual judges the presence of a cause when informed of the presence of one of its effects, and whether and to what degree the individual becomes less confident in the cause when informed that another cause of the effect was present. Kelly [1972] called this inference *discounting*. Several researchers ([Jones, 1979], [Quattrone, 1982], [Einhorn and Hogarth, 1982], [McClure, 1989]) have argued that studies indicate that in certain situations people discount less than is warranted. On the other hand, arguments that people discount more than is warranted also have a long history (See, [Kanouse, 1972], and [Nisbett and Ross, 1980]). In many of the discounting studies, individuals were asked to state their feelings about the presence of a particular cause when informed another cause was present. For example, a classic finding is that subjects who read an essay defending Fidel Castro's regime in Cuba ascribe a pro-Castro attitude to the essay writer even when informed that the writer was instructed to take a pro-Castro stance. Researchers interpreted these results as indicative of underdiscounting. Morris and Larrick [1995] argue that the problem in these studies is that the researchers assume that subjects believe a cause is sufficient for an effect when they do not. Morris and Larrick [1995] repeated the Castro studies, but used subjective probability testing instead of assuming, for example, that the subject believes an individual will always write a pro-Castro essay whenever told to do so (They found that subjects only felt it was highly probable this would happen). In *subjective probability* testing, the subject is given a problem that can be solved using applications of Bayes' rule, and then asked to judge the probability in each component of the problem along with the solution. A solution (called the *normative* solution) is then obtained by applying Bayes' rule to the subject's subjective probabilities. Finally, that solution is compared to the subject's solution. When they replaced deterministic relationships by probabilistic ones, Morris and Larrick [1995] found that subjects discounted about *normatively*. This means, when reasoning with multiple causes, they approximately reached the same judgments as algorithms for doing inference in Bayesian networks.

## 3.2 LEARNING CAUSES

Next we present an argument that humans learn causal structure by observing the same independencies and dependencies used by certain algorithms for learning the structure of a Bayesian network. Indeed, we argue that the very notion of causality develops from the observation of these independencies/dependencies.

### 3.2.1 AN ALGORITHM FOR LEARNING CAUSAL STRUCTURE

Part of our intuition concerning cause-effect relationships is that an effect cannot precede one of its causes in time. Therefore, if we have the benefit of a time ordering of the variables, we can learn causal influences with an algorithm based on the theorem that follows. First we need some notation. When two variables are independent conditional on some subset of variables, $\mathcal{S}$ (possibly empty), we denote this by $I(X, Y|\mathcal{S})$.

**Theorem 1** *Suppose a probability distribution $\mathcal{P}$ satisfies the Markov and faithfulness conditions for some DAG $\mathcal{D}$ whose nodes are the elements in $\mathcal{U}$. Then there is a directed path from $X$ to $Y$ in $\mathcal{D}$ if there is a third variable $Z$ and a set of variables $\mathcal{S}_{XY}$, such that $Z$ and all of the elements in $\mathcal{S}_{XY}$ are not descendents of $X$, satisfying the following:*

1. $\neg I(Z, Y|\mathcal{S}_{XY})$

2. $I(Z, Y|\mathcal{S}_{XY} \cup \{X\})$.

If we have a time ordering of the variables, any variable preceding another in time could not be a descendent of that variable. Therefore, the following algorithm follows from Theorem 1. Given the assumption that the observed variables are a subset of variables in a DAG $\mathcal{D}$ for which $\mathcal{P}$ satisfies the Markov and faithfulness conditions, and given as input independencies and dependencies among the observed variables, the algorithm produces links which represent directed paths in $\mathcal{D}$.

### Algorithm I

**order** the variables according to time;
**for** each pair of variables, $X$ and $Y$ such that $X$ precedes $Y$ **do**
  **search** for a set $\mathcal{S}_{XY}$ and a variable $Z$ such that $Z$ and all variables in $\mathcal{S}_{XY}$ precede $X$ **and**
    1. $\neg I(Z, Y|\mathcal{S}_{XY})$
    2. $I(Z, Y|\mathcal{S}_{XY} \cup \{X\})$
  **end**;
  **if** such a couple is found **then**
    create a link $X \rightarrow Y$, which means $X$ has a causal influence on $Y$
  **end**
**end**;

Similar, more detailed algorithms, which do not require a time ordering of the variables, appear in [Pearl and Verma, 1991], [Spirtes et al, 1993], and [Glymour, 1996].

3.2.2 WHAT IS A CAUSE?

Rather than offering an explicit definition of causality, the probabilistic causal graph model assumes that the probability distribution satisfies certain properties (the Markov and faithfulness conditions) we intuitively feel hold for causal relationships. Implicitly, a cause is then defined to be any directed link learned by algorithms like the preceding one. To discover a cause, the algorithm must be given a set of variables. But how do we know what these variables are? Consider the following example taken from [Spirtes et al, 1993, p. 42].

> If $C$ is the event of striking a match, and $A$ is the event of the match catching on fire, and no other events are considered, then $C$ is a direct cause of $A$. If, however, we added $B$; the sulfur on the matchtip achieved sufficient heat to combine with the oxygen, then we could no longer say that $C$ directly caused $A$, but rather $C$ directly caused $B$ and $B$ directly caused $A$. Accordingly, we say that $B$ is a causal mediary between $C$ and $A$ if $C$ causes $B$ and $B$ causes $A$.

Clearly, we can add more causal mediaries. For example, we could add the event $D$ that the match tip is abraded by a rough surface. $C$ would then cause $D$, which would cause $B$, etc. We could go much further and describe the chemical reaction that occurs when sulfur combines with oxygen. Indeed, it seems we can conceive of a continuum of events in any causal description of a process. We see then that the set of observable variables is observer dependent. Apparently, an individual, given a myriad of sensory input, selectively records discernible events and develops cause-effect relationships among them. Therefore, for the purpose of modeling human thought, rather than assuming there is a set of causally related variables out there, it seems more appropriate to only assume that each individual develops a set of 'causal' relationships among variables, which are specific to the individual (although many are shared), and the individual reconstructs these relationships on-the-fly when reasoning.

What is this relationship among variables that the notion of causality embodies? It seems reasonable to assume that most human knowledge derives from statistical observations, and that therefore a causal relationship must recapitulate some statistical observation among variables. Should we look at the adult to learn what this statistical observation might be? As Piaget and Inhelder [1969, p. 157] note, 'Adult thought might seem to provide a preestablished model, but the child does not understand adult thought until he has reconstructed it, and thought is itself the result of an evolution carried on by several generations, each of which has gone through childhood.' The intellectual concept of causality has been developed through many generations and knowledge of many (if not most) cause-effect relationship are passed on to individuals by previous generations. Piaget and Inhelder [1969, p. ix] note further 'While the adult educates the child by means of multiple social transmissions, every adult, even if he is a creative genius, begins as a small child.' So we will look to the small child, indeed to the infant, for the genesis of the concept of causality. We will discuss results of studies by Piaget. We will show how these results can lead us to a definition of causality as a statistical relationship among an individual's observed variables.

3.2.3 THE GENESIS OF THE CONCEPT OF CAUSALITY

Piaget [1952, 1954, 1966] established a theory of the development of sensori-motor intelligence in infants from birth until about age two. He distinguished six stages within the sensori-motor period. Our purpose here is not to recount these stages, but rather to discuss some observations Piaget made in several stages, which might shed light on what observed relationships the concept of causality recapitulates.

Piaget argues that the mechanism of learning 'consists in assimilation; meaning that reality data are treated or modified in such a way as to become incorporated into the structure...According to this view, the organizing activity of the subject must be considered just as important as the connections inherent in the external stimuli.'- [Piaget and Inhelder, 1969, p. 5]. We will investigate how the infant organizes external stimuli into cause-effect relationships.

The third sensori-motor stage goes from about the age of four months to nine months. Here is a description of what Piaget observed in infants in this stage (taken from [Drescher, 1991, p. 27]):

> Secondary circular reactions are characteristic of third stage behavior; these consist of the repetition of actions in order to reproduce fortuitously-discovered effects on objects. For example:
>
> - The infant's hand hits a hanging toy. The infant sees it bob about, then repeats the gesture several times, later applying it to other objects as well, developing a striking schema for striking.
>
> - The infant pulls a string hanging from the bassinet hood and notices a toy, also connected to the hood, shakes in response. The infant again grasps and pulls the string, already watching the toy rather than the string. Again, the spatial and causal nature of the connection between the objects is not well understood; the infant will generalize the gesture to inappropriate situations.

Piaget and Inhelder [1969, p. 10] discuss these inappropriate situations:

> Later you need only hang a new toy from the top of the cradle for the child to look for the cord, which constitutes the beginning of a differentiation between means and end. In the days that follow, when you swing an object from a pole two yards from the crib, and even when you produce unexpected and mechanical sounds behind a screen, after these sights or sounds have ceased the child will look for and pull the magic cord. Although the child's actions seem to reflect a sort of magical belief in causality without any material connection, his use of the same means to try to achieve different ends indicates that he is on the threshold of intelligence.

Piaget and Inhelder [1969, p. 18] note that 'this early notion of causality may be called magical phenomenalist; "phenomenalist"; because the phenomenal contiguity of two events is sufficient to make them appear causally related.' At this point, the notion of causality in the infant's model entails a primitive cause-effect relationship between actions and results. For example if $Z=$'pull string hanging from bassinet hood' and $Y=$'toy shakes', the infant's model contains the causal relationship $Z \to Y$. The infant extends this relationship to believe there may be an arrow from $Z$ to other desired results even when they were not preceded by $Z$. Drescher [1991, p. 28] states that the 'causal nature of the connection between the objects is not well understood.' Since our goal here is to determine what relationships the concept of causality recapitulates, we do not want to assume there is a 'causal nature of the connection' that is actually out there. Rather we could say that at this stage an infant is only capable of forming two-variable relationships. The infant cannot see how a third variable may enter into the relationship between any two. For example, the infant cannot develop the notion that the hand is moving the bassinet hood, which in turn makes the toy shake.

Although there are advances in the fourth stage (about age nine months to one year), the infant's model still only includes two variables relationships during this stage. It is in the fifth stage (commencing at about one year of age) that the infant sees a bigger picture. Here is an account by [Drescher, 1991, p. 34] of what can happen in this stage:

> You may recall that some secondary circular reactions involved influencing one object by pulling another connected to the first by a string. But that effect was discovered entirely by accident, and, with no appreciation of the physical connection. During the present stage, the infant wishing to influence a remote object learns to search for an attached string, visually tracing the path of connection.

Piaget and Inhelder [1969, p. 19] describe this fifth stage behavior as follows:

> In the behavior patterns of the support, the string, and the stick, for example, it is clear that the movements of the rug, the string, or the stick are believed to influence those of the subject (independently of the author of the displacement).

If we let $Z=$'pull string hanging from bassinet hood', $X=$'bassinet hood moves', and $Y=$'toy shakes', at this stage the infant develops the relationship that $Z$ is connected to $Y$ through $X$. At this point, the infant's model entails that $Z$ and $Y$ are dependent, but that $X$ is a causal mediary and that they are independent given $X$. Using our previous notation, this relationship is expressed as follows:

$$\neg I(Z, Y) \qquad I(Z, Y|X) \qquad (1)$$

The fifth stage infant shows no signs of mentally simulating the relationship between objects and learning from the simulation instead of from actual experimentation. So it can only form causal relationships by repeated experiments. Furthermore, although it seems to recognize the conditional independence, it does not seem to recognize a causal relationship between $X$ and $Y$ that is merely learned via $Z$. Because it only learns from actual experiments, the third variable is always part of the relationship. This changes in the sixth stage. Piaget and Inhelder [1969, p. 11] describe this stage as follows:

> Finally, a sixth stage marks the end of the sensori-motor period and the transition to the following period. In this stage the child becomes capable of finding new means not only be external or physical groping but also by internalized combinations that culminate in sudden comprehension or *insight*.

Drescher [1991, p. 35] gives the following example of what can happen at this stage:

An infant who reaches the sixth stage without happening to have learned about (say) using a stick may invent that behavior (in response to a problem that requires it) quite suddenly.

It is in the sixth stage that the infant recognizes an object will move as long as something hits it (e.g. the stick); that there need by no specific learned sequence of events. Therefore, at this point the infant recognizes the movement of the bassinet hood as a *cause* of the toy shaking, and that the toy will shake if the hood is moved by any means whatsoever.

The argument here is not that the two-year-old child has causal notions like those of the adult. Rather that they are as described by Piaget and Inhelder [1969, p. 13]:

> It organizes reality by constructing the broad categories of action which are the schemes of the permanent object, space, time, and causality, substructures of the notions that will later correspond to them. None of these categories is given at the outset, and the child's initial universe is entirely centered on his own body and action in an egocentrism as total as it is unconscious (for lack of consciousness of the self). In the course of the first eighteen months, however, there occurs a kind of Copernican revolution, or, more simply, a kind of general decentering process whereby the child eventually comes to regard himself as an object among others in a universe that is made up of permanent objects and in which there is at work a causality that is both localized in space and objectified in things.

Piaget and Inhelder [1969, p. 90] feel that these early notions are the foundations of the concepts developed later in life:

> The roots of logic are to be sought in the general coordination of actions (including verbal behavior) beginning with the sensori-motor level, whose schemes are of fundamental importance. This schematism continues thereafter to develop and to structure thought, even verbal thought, in terms of the progress of actions, until the formation of the logico-mathematical operations.

Piaget found that the development of the intellectual notion of causality mirrors the development of the infant's notion. This is discussed in Piaget and Inhelder [1969, p. 110]:

> The stars "were born when we were born," says the boy of six, "because before that there was no need for sunlight." ... Interestingly enough, this precausality is close to the initial sensori-motor forms of causality, which we called "magical-phenomenalist" in Chapter 1. Like those, it results from a systematic assimilation of physical processes to the child's own action, an assimilation which sometimes leads to quasi-magical attitudes (for instance, many subjects between four and six believe that the moon follows them....) But, just as sensori-motor precausality makes way (after Stages 4 to 6 of infancy) for an objectified and spacialized causality, so representative precausality, which is essentially an assimilation to actions, is gradually, at the level of concrete operations, transformed into a rational causality by assimilation no longer to the child's own action in their egocentric orientation but to the operations as general coordination of actions.

In the period of concrete operations (between the ages of seven and eleven), the child develops the adult concept of causality. According to Piaget, that concept has its foundations in the notion of objective causality developed at the end of the sensori-motor period.

In summary, we have offered the hypothesis that the concept of causality develops in the individual, starting in infancy, through the observation of statistical relationships among variables and we have given supportive evidence for that hypothesis. But what of the properties of *actual causal relationships* that a statistical explanation may not seem to address? For example, consider the child who moves the toy by pulling the rug on which it is situated. We said that the child develops the causal relationship that the moving rug causes the toy to move. An adult, in particularly a physicist, would have a far more detailed explanation. For example, the explanation might say that the toy is sufficiently massive to cause a downward force on the rug so that the rug does not slide from underneath the toy, etc. However, such an explanation is not unlike that of the child's; it simply contains more variables based on the adult's keener observations and having already developed the intellectual concept of causality. Piaget and Inhelder [1969, p. 19] note that even the stage five infant requires physical contact between the toy and rug to infer causality:

> If the object is placed beside the rug and not on it, the child at Stage 5 will not pull the supporting object, whereas the child at Stage 3 or even 4 who has been trained to make use of the supporting object will still pull the rug even if the object no longer maintains with it the spatial relationship "placed upon."

This physical contact is a necessary component to the child forming the causal link, but it is not the mechanism by which the link develops. The hypothesis here is that this mechanism is the observed statistical relationships among the variables. Concern over *actual causal relationships* is not pertinent in a psychological investigation into the genesis of the concept of causality because that concept is part of the human model; not part of reality itself. As Kant [1787] noted long ago, we cannot truly gain access to what is 'out there'. What is pertinent is how humans assimilate reality into the concept of causality. We are hypothesizing that this concept developed to describe the observed statistical relationship among variables shown in this section.

### 3.2.4   A Definition of Causality

Considerations of the Markov and faithfulness conditions led Pearl and Verma [1991] to the following definition of causality.

**Definition 2** *A variable $X$ has a causal influence on a variable $Y$ if there is a third variable $Z$ and a context $S_{XY}$, both occurring before $X$ such that:*

1. $\neg I(Z, Y | S_{XY})$

2. $I(Z, Y | S_{XY} \cup \{X\})$

By a context $S_{XY}$, they mean any set (including the empty set) of instantiated variables. Note that this definition is exactly the conditions used in Algorithm I to deduce directed paths in a Bayesian network.

Given that the infant is always in some context, our expression (Expression 1), which summarizes the infant's perceived relationships among variables, is identical to Pearl's definition. We conjecture that the intuition for the Markov and faithfulness conditions also develop when the infant is first capable of modeling with three variables, and we instead offer Definition 2 as a definition of causality based on our argument that the concept developed as a recapitulation of the statistical relationships in this definition. Since the variables are specific to an individual's observations, this is a subjective definition of causality not unlike the subjective definition of probability. Indeed, since it is based on statistical relationships, one could say it is in terms of that definition. According to this view, there are no objective causes as such. Bertrand Russell [1913] long ago noted that causation played no role in physics and wanted to eliminate the word from science. Similarly, Karl Pearson [1911] wanted it removed from statistics. Whether this would be appropriate for these disciplines is another issue. However, the concept is important in psychology and artificial intelligence because humans model the exterior in terms of causation. We have suggested that the genesis of the concept lies in the statistical relationship shown above. If this so, for the purposes of these disciplines, the statistical definition would be accurate. This definition simplifies the task of the researcher in artificial intelligence as we need not engage in metaphysical wrangling about causality.

## 4   Future Research

A considerable reason for understanding how humans reason with and learn causes is that this understanding may give us clues as how to create a lifelike learning and reasoning system. Current algorithms for learning causal structure require the set of variables and probability distribution as input; they then produce the links. However, this does not appear to be the way humans learn. Furthermore, such a method could not be used by an autonomous agent that must learn, react, and make decisions in a complex, dynamic environment. As noted by Pearl, 'human beings seem to learn in bursts, but how can we create artificial systems that learn this way?' — [private correspondence, 1995]. Stewart and Peregov [1983] have already used catastrophe theory to model sudden belief changes. By studying human subjects, we can gain further insight into sudden learning. We plan to test the accuracy of the hypotheses forwarded in this paper and to learn more about how humans learn causes. Next we briefly describe our plans.

As discussed in Section 3.1, studies by Morris and Larrick [1995] substantiate that humans reach normative judgments when reasoning with common causes. We plan to do more subjective probability testing to see if humans reach normative judgments when reasoning with indirect causes and with multiple effects. Subjective probability testing only examines the *results* of reasoning, not the *process*. That is, even if humans reach normative judgments, it does not mean they reach them by traversing links as suggested by Pearl. We will use *prime-probe concept pairing reaction time* studies to test the cognitive representation of cause-effect relationships and whether humans reason by traversing links.

As to investigating how causes are learned, previous studies ([Heider, 1944], [Kelly, 1967]) indicate that humans learn causes to satisfy a need for prediction and control of their environment. Putting people into an artificial environment, with a large number of cues, and forcing them to predict and control the environment should produce the same types of causal reasoning that occurs naturally. One option is some sort of computer game. Occasionally, the game would be interrupted, and subjects would be asked to perform reaction time tasks.

The goal would be to see how and when subjects learned new causal relationship, in particular to see how subjects learn in bursts.